\documentclass[10pt,twocolumn,letterpaper]{article}

\usepackage[pagenumbers]{cvpr} 

\usepackage[accsupp]{axessibility}
\usepackage{booktabs}
\usepackage{multirow}
\usepackage{amsmath}
\usepackage{amssymb}
\usepackage{graphicx}
\usepackage{xspace}
\usepackage{tikz}
\usetikzlibrary{positioning}
\usepackage{pgfplots}
\pgfplotsset{compat=1.18}
\usepackage{subcaption}

\newcommand{\SenBenMovies}{157}
\newcommand{\SenBenFrames}{13{,}999}

\newcommand{\SenBenTestMovies}{31}
\newcommand{\SenBenTestFrames}{2{,}000}

\newcommand{\SenBenObjects}{25}
\newcommand{\SenBenAttributes}{28}
\newcommand{\SenBenPredicates}{14}
\newcommand{\SenBenTags}{16}
\newcommand{\SenBenCategories}{5}

\newcommand{\senben}{SenBen\xspace}
\newcommand{\fl}{Florence-2\xspace}

\newcommand{\SBR}{\mathrm{R}_{\mathrm{SB}}}
\newcommand{\SBP}{\mathrm{P}_{\mathrm{SB}}}
\newcommand{\SBF}{\text{F1}_{\mathrm{SB}}}

\usepackage{capt-of}
\makeatletter
\g@addto@macro\@maketitle{%
  \par\vspace{6pt}%
  \centering
  \includegraphics[width=0.8\textwidth]{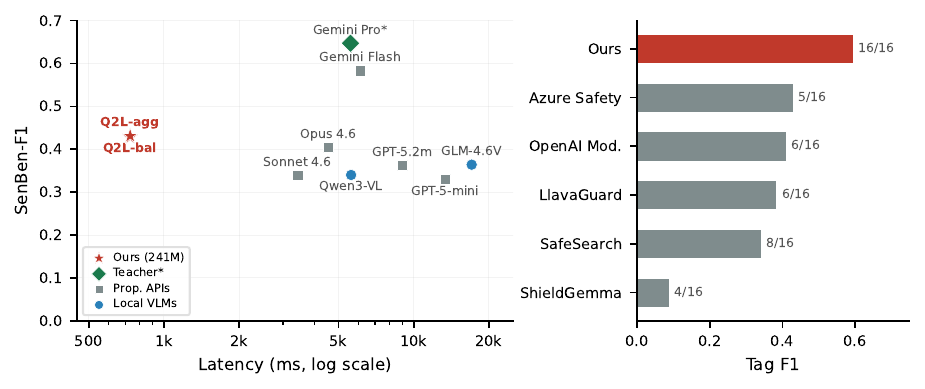}\par
  \vspace{2pt}%
  \captionof{figure}{Our model (using only 1.2\,GB VRAM) achieves the best trade-off between speed, accuracy, and tag coverage among all evaluated models. \textbf{Left}: Latency \vs SenBen F1 on 2{,}000 test frames; our model (stars) is $7.6\times$ faster than the next local VLM and competitive with proprietary APIs at zero inference cost. *Gemini~Pro generated initial labels; human-reviewed and corrected. \textbf{Right}: Tag detection F1 against commercial safety classifiers; our model is the only one covering all 16 sensitivity tags.}%
  \label{fig:teaser}%
  \vspace{6pt}%
}
\makeatother

\definecolor{cvprblue}{rgb}{0.21,0.49,0.74}
\usepackage[breaklinks,colorlinks,allcolors=cvprblue]{hyperref}

\def\paperID{3} 
\def\confName{CVPR}
\def\confYear{2026}

\title{SenBen: Sensitive Scene Graphs for Explainable Content Moderation\thanks{Code and data: \url{https://github.com/fcakyon/senben}}}

\author{
Fatih Cagatay Akyon$^{1,2}$ \qquad Alptekin Temizel$^{1}$\\[6pt]
$^1$Graduate School of Informatics, METU, Turkiye\\
$^2$Ultralytics, Inc., United States
}

\begin{document}
\maketitle
\begin{abstract}
Content moderation systems classify images as safe or unsafe but lack spatial grounding and interpretability: they cannot explain \emph{what} sensitive behavior was detected, \emph{who} is involved, or \emph{where} it occurs.
We introduce the Sensitive Benchmark (\senben), the first large-scale scene graph benchmark for sensitive content, comprising \SenBenFrames{} frames from \SenBenMovies{} movies annotated with Visual Genome-style scene graphs (\SenBenObjects{} object classes, \SenBenAttributes{} attributes including affective states such as \emph{pain}, \emph{fear}, \emph{aggression}, and \emph{distress}, \SenBenPredicates{} predicates) and \SenBenTags{} sensitivity tags across \SenBenCategories{} categories.
We distill a frontier VLM into a compact 241M student model using a multi-task recipe that addresses vocabulary imbalance in autoregressive scene graph generation through suffix-based object identity, Vocabulary-Aware Recall (VAR) Loss, and a decoupled Query2Label tag head with asymmetric loss, yielding a +6.4 percentage point improvement in SenBen Recall over standard cross-entropy training.
On grounded scene graph metrics, our student model outperforms all evaluated VLMs except Gemini models and all commercial safety APIs, while achieving the highest object detection and captioning scores across all models, at $7.6\times$ faster inference and 16$\times$ less GPU memory.
\end{abstract}

\section{Introduction}
\label{sec:intro}

Scaling content moderation necessitates automation, as manual review is not only prohibitively slow and expensive but also psychologically damaging to human moderators~\cite{steiger2021moderator}.
However, current automated approaches (convolutional classifiers, vision transformers, and commercial APIs) produce opaque labels such as ``unsafe'' or ``sexual'' without explaining \emph{what} behavior was detected or \emph{where} it occurs in the image.
This lack of interpretability prevents auditing, cultural adaptation across platforms with different content policies, and meaningful human oversight.

Scene graphs offer an alternative by offering structured representations where objects with attributes are connected by predicates.For example, a scene graph stating \texttt{male:aggression} $\xrightarrow{\text{hitting}}$ \texttt{female:distress} provides machine-readable, queryable, and spatially-grounded evidence.
Unlike post-hoc saliency maps that produce noisy activations on decoder-based VLMs, or natural language rationales~\cite{helff2025llavaguard} that lack spatial grounding, scene graphs are inherently interpretable: the output \emph{is} the explanation.
Crucially, different platforms can apply different predicate and attribute thresholds without retraining; a scene graph is a structured intermediate representation that decouples detection from policy.

No existing benchmark provides large-scale Visual Genome-style scene graph annotations for sensitive content.
The closest work, USD~\cite{zhang2025usd}, provides 1,300 images with subject-verb-object triples, entity attributes, and six unsafe categories.
Existing moderation datasets (LSPD~\cite{phan2022lspd}, NudeNet~\cite{nudenet2019}) offer at most body-part localization without relational scene graph structure, and are effectively saturated, with convolution-based classifiers reaching 0.95+ F1~\cite{akyon2023nudity}.
Meanwhile, attention-based explainability methods (GradCAM, cross-attention rollout) produce noisy, spatially uncorrelated activations on decoder-based VLMs, because pretrained vision encoders lack sensitive-content priors and autoregressive decoding dilutes spatial signal across tokens.
Even recent SGG debiasing methods~\cite{peddi2025impartail, yoon2025rasgg} remain in the two-stage fixed-vocabulary paradigm, generating neither attributes, captions, nor sensitivity tags.
These gaps motivate structured scene graph annotations as the path to explainable moderation.

Movies are an established academic data source for behavioral understanding~\cite{huang2020movienet, rohrbach2017lsmdc, vicol2018moviegraphs, marszalek2009hollywood2, schedl2015vsd}, offering rich acted affective behaviors: the expression attributes in \senben (\emph{pleasure}, \emph{pain}, \emph{fear}, \emph{aggression}, \emph{distress}) directly capture affective dimensions relevant to behavioral analysis.
The MECD dataset~\cite{mecd2024kaggle} provides timestamp annotations for 1,734 movies across 30 sensitivity tags across 6 categories, enabling scalable frame extraction without manual screening.
Following the Visual Genome paradigm, our task treats each extracted frame as an independent image for scene graph generation: grounded sensitive-content understanding is unsolved even at single-image level, and movies provide the diversity and realism that static image collections lack.
We distill a frontier VLM into a compact student via knowledge distillation~\cite{hinton2015distilling} through structured pseudo-labeling, producing a model that runs locally on any GPU (${\sim}$1.2\,GB VRAM) at 733\,ms per frame, $7.6{\times}$ faster than the next local VLM and at zero per-frame cost compared to commercial APIs (Figure~\ref{fig:teaser}).

Our contributions are:
\begin{itemize}
    \item \textbf{SenBen: A Grounded Benchmark for Explainable Moderation}: We introduce the first large-scale scene graph benchmark specifically for sensitive content, comprising \SenBenFrames{} frames from \SenBenMovies{} movies. The dataset features Visual Genome-style annotations, \SenBenTags{} sensitivity tags in \SenBenCategories{} categories, and a recall-focused composite metric (SenBen-Score).
    \item \textbf{Multi-Task Knowledge Distillation with Vocabulary-Aware Optimization}: We propose a novel training recipe to distill a frontier VLM into a compact 241M student model. Our approach combines suffix-based object identity, Vocabulary-Aware Recall Loss (VAR), and a decoupled Query2Label (Q2L) tag head with asymmetric loss. This recipe yields a +6.4pp improvement in SenBen Recall over cross-entropy training.
    \item \textbf{High-Efficiency Local Inference}: We demonstrate that our student model outperforms all evaluated VLMs except Gemini models and commercial safety APIs on grounded scene graph metrics. It achieves the highest object detection and captioning scores, at $7.6{\times}$ faster inference and requiring 16$\times$ less GPU memory (1.2 GB VRAM) than the next best performing local VLM.
\end{itemize}

\section{Related Work}
\label{sec:related}

\noindent\textbf{Content moderation datasets and APIs.}
Existing datasets span nudity, violence, and substance use but lack relational scene graph structure: LSPD~\cite{phan2022lspd} and NPDI~\cite{npdi_2k_dataset} (pornography), Violent Scenes Dataset~\cite{schedl2015vsd} (violence in movies), NudeNet~\cite{nudenet2019} (nudity detection), and substance use detection~\cite{substance_use_dataset} (binary drug/not; \senben further distinguishes legal \vs illegal substances with action predicates such as \emph{snorting} and \emph{injecting}).
Convolution-based classifiers already reach 0.95{+} F1 on these benchmarks~\cite{akyon2023nudity}, leaving little room for improvement within the binary-label paradigm.
Commercial APIs (OpenAI Moderation~\cite{openai_moderation}, Azure Content Safety~\cite{azure_content_safety}, Google SafeSearch~\cite{google_safesearch}) cover 4--11 categories with severity scores but provide no spatial grounding.
Safety classifiers such as ShieldGemma~2~\cite{zeng2025shieldgemma2} (4B) and LlavaGuard~1.2~\cite{helff2025llavaguard} (7B) add policy-based reasoning but still output flat decisions.
UnsafeBench~\cite{qu2025unsafebench} benchmarks 11 categories on real and AI-generated images; none of the evaluated classifiers provide spatial grounding.
USD~\cite{zhang2025usd} is the closest prior work: it applies scene graphs to NSFW detection on 1,300 images with flat (subject, verb, object) triples and binary safe/unsafe classification.
USD provides manual annotations with strong inter-annotator agreement ($\kappa{=}0.94$) and evaluates open-scenario transferability.
\senben extends this to 13,999 real movie frames at ${\sim}$11$\times$ scale, with richer Visual Genome-style annotations (28 attributes \vs 9), structured generation as the task, and 16 fine-grained tags \vs 6 scenario types.
The two are complementary: USD addresses text-to-image safety, \senben addresses media content moderation.\\
\noindent\textbf{Scene graph generation.}
Classical SGG methods Neural Motifs~\cite{zellers2018neural}, TDE~\cite{tang2020unbiased}) and recent debiasing work, IMPARTAIL~\cite{peddi2025impartail}, use fixed-vocabulary detectors with parallel predicate classifiers, generating neither attributes, captions, nor sensitivity tags.
Autoregressive approaches encode visual structures as text: Pix2Seq~\cite{pix2seq} for detection, FactualSceneGraph~\cite{factual_scene_graph} for triplet generation with suffix-based object identity.
\fl~\cite{xiao2024florence2} unifies detection, captioning, and grounding via inline \texttt{<loc>} spatial tokens in a 230M encoder-decoder.
R1-SGG~\cite{chen2025r1sgg} refines MLLM scene graphs via GRPO reinforcement learning on 2B--7B models, using post-hoc reward matching rather than in-loss differentiable training.
USD uses a multi-stage pipeline (OpenSeeD $\rightarrow$ BLIP $\rightarrow$ BERT classifier).
Our approach generates full scene graphs end-to-end in a single compact model via knowledge distillation~\cite{hinton2015distilling}: a frontier VLM teacher generates structured pseudo-labels that the student learns to reproduce, analogous to Distil-Whisper~\cite{gandhi2023distilwhisper} for speech recognition.\\
\noindent\textbf{Affective behavior in movies.}
Movies are a natural source of acted affective behaviors, and several datasets annotate them at various granularities: MovieGraphs~\cite{vicol2018moviegraphs} provides social-situation graphs with character interactions and emotions, VSD~\cite{schedl2015vsd} annotates violent segments, and the ABAW workshop series~\cite{kollias2023abaw} drives affective computing on in-the-wild video.
\senben connects to this line of work through its expression attributes (\emph{pleasure}, \emph{pain}, \emph{fear}, \emph{aggression}, \emph{distress}), which capture affective dimensions within a spatially-grounded scene graph structure.
Unlike flat emotion labels, scene graphs provide spatial grounding for \emph{who} exhibits which affective state and \emph{what} triggers it.\\
\noindent\textbf{Loss functions for imbalanced generation.}
Focal Loss~\cite{lin2017focal} and Asymmetric Loss (ASL)~\cite{ridnik2021asymmetric} address class imbalance in classification.
Recall Loss~\cite{tian2022recall} weights segmentation classes by current recall; Skeleton Recall Loss~\cite{skeleton_recall_loss} adds an additive soft recall term for thin structures.
SGG-specific debiasing losses, PPDL~\cite{li2022ppdl}, CDL~\cite{lyu2022fine}, RA-SGG~\cite{yoon2025rasgg} (inverse propensity scoring), and IMPARTAIL~\cite{peddi2025impartail} (progressive masking), operate on parallel predicate classification heads, not autoregressive token sequences.
For order-invariant training, OaXE~\cite{du2021oaxe} uses Hungarian matching for non-autoregressive models, and $\sigma$-GPTs~\cite{pannatier2024sigmagpts} train on randomly shuffled sequences.
None address vocabulary-level imbalance in autoregressive text generation, where sensitive tokens are diluted among coordinates and punctuation, the gap our VAR Loss fills.
\section{Method}
\label{sec:method}
\subsection{SenBen Dataset}
\label{sec:dataset}

Initial data extraction was performed using sensitivity timestamps from the MECD dataset~\cite{mecd2024kaggle} , which provides annotations for 1,734 movies across 30 tags in 6 categories.
We selected the 16 visually detectable tags (excluding audio-based language categories) and organize them into \SenBenCategories{} categories: \emph{immodesty} (immodesty, nudity, nudity art, nudity implied), \emph{sexual} (sexually suggestive, kissing, sex implied, sexual activity), \emph{violence} (violence, gore), \emph{substances} (drugs legal, drugs illegal), and \emph{other} (bodily functions, vulgar gestures, medical graphic, medical procedures).\\
\noindent\textbf{Construction pipeline.}
For each movie, we detected shot boundaries using PySceneDetect, extracted representative frames from shots overlapping with MECD sensitivity windows ($\pm$30s padding), and labeled each frame using Gemini~3~Pro~\cite{gemini3_2025} with thinking mode (level \texttt{high}, temperature 0.1).
The prompt enforces a forensic scanning protocol with structured vocabulary constraints.
These initial labels were then refined through human review via a custom web interface, including vocabulary normalization, tag validation, and scene graph correction.
The final dataset was stratified into train/val/test splits (65/35 mature \vs other ratings).
This pipeline constitutes knowledge distillation~\cite{hinton2015distilling} via structured pseudo-labeling with human correction: Gemini~3~Pro (teacher) generates initial scene graph annotations, which are then human-reviewed and corrected before 241M \fl student model learns to reproduce them.\\
\noindent\textbf{Statistics.}
Table~\ref{tab:split-overview} summarizes the split statistics for \senben, which contains \SenBenFrames{} frames from \SenBenMovies{} movies with a controlled 50/50 balance between sensitive/general content.
The vocabulary comprises \SenBenObjects{} object classes (e.g., persons, weapons, substances), \SenBenAttributes{} attributes organized in 6 groups (body state, pose, clothing condition, exposure, gore, and expression), and \SenBenPredicates{} predicates in 5 groups (sexual, violence, interaction, substance, and gesture).
Each frame is annotated as a JSON scene graph containing sensitivity tags, a natural language caption, objects with bounding boxes ($[y_\text{min}, x_\text{min}, y_\text{max}, x_\text{max}]$ normalized to a $[0,1000]$) scale, alongside their attributes, and predicate triplets.
Table~\ref{tab:tag-distribution} shows the per-tag distribution across splits, highlighting that while violence and immodesty categories dominate the dataset, categories like vulgar gestures and bodily functions form a distinct long tail.
Total annotation cost was under $\$250$ ($~\$0.02$/frame).
Movies span ratings R (87), PG-13 (44), TV-MA (15), and PG (7), with PG-13 contributing nearly equal sensitive content representation.
Split curation was guided by six bias metrics (HHI, nPMI, DISC, REVISE, log-odds, statistical lift) to ensure robust evaluation.\\
\begin{table}[t]
\centering
\caption{Dataset split statistics of \senben. Annotation density is averaged per frame.}
\label{tab:split-overview}
\setlength{\tabcolsep}{4pt}
\small
\begin{tabular}{@{}lrrrrrrr@{}}
\toprule
Split & Frames & Movies & Sens. & Gen. & Obj/fr & Attr/fr & Pred/fr \\
\midrule
Train &  9{,}999 & 95  & 4{,}999 & 5{,}000 & 3.42 & 8.09 & 2.50 \\
Val   &  2{,}000 & 31  & 1{,}000 & 1{,}000 & 3.39 & 7.95 & 2.50 \\
Test  &  2{,}000 & 31  &    989  & 1{,}011 & 3.41 & 8.04 & 2.52 \\
\midrule
Total & 13{,}999 & 157 & 6{,}988 & 7{,}011 & 3.41 & 8.06 & 2.51 \\
\bottomrule
\end{tabular}
\end{table}
\begin{table}[t]
\centering
\caption{Per-tag frame counts across splits. Frames may carry multiple tags (avg 1.3 per sensitive frame).}
\label{tab:tag-distribution}
\small
\begin{tabular}{@{}llrrrr@{}}
\toprule
Category & Tag & Train & Val & Test & Total \\
\midrule
\multirow{4}{*}{Immodesty}
  & immodesty        &   734 & 177 & 155 & 1,066 \\
  & nudity           &   370 &  62 &  88 &   520 \\
  & nudity art       &   143 &  25 &  31 &   199 \\
  & nudity implied   &   371 &  61 &  68 &   500 \\
\midrule
\multirow{4}{*}{Sexual}
  & sex.\ suggestive &   579 & 104 & 111 &   794 \\
  & kissing          &   248 &  47 &  49 &   344 \\
  & sex implied      &   182 &  23 &  28 &   233 \\
  & sexual activity  &   216 &  29 &  44 &   289 \\
\midrule
\multirow{2}{*}{Violence}
  & violence         & 1,792 & 374 & 355 & 2,521 \\
  & gore             &   315 &  54 &  69 &   438 \\
\midrule
\multirow{2}{*}{Substances}
  & drugs legal      &   985 & 216 & 201 & 1,402 \\
  & drugs illegal    &   243 &  40 &  47 &   330 \\
\midrule
\multirow{4}{*}{Other}
  & bodily functions &   100 &  11 &  20 &   131 \\
  & vulgar gestures  &    63 &   7 &   8 &    78 \\
  & medical graphic  &   196 &  23 &  29 &   248 \\
  & medical proc.    &   188 &  20 &  29 &   237 \\
\bottomrule
\end{tabular}
\end{table}
\noindent\textbf{Ethics.}
Movies are ethically-sourced professional content with established MPAA ratings.
No personally identifiable information is included beyond actors in commercially released films.
The dataset, model weights, and code will be released under gated access with a research-only license requiring institutional affiliation and stated academic purpose.

\subsection{SenBen-Score}
\label{sec:metric}

We define a recall-focused composite metric that evaluates four scene graph components independently, then averages across sensitivity categories.

For each category $c \in \mathcal{C}$, where $\mathcal{C} = \{\text{immodesty}, \text{sexual}, \text{violence}, \text{substances}, \text{other}\}$, we compute the per-class macro-averaged mean recall for tags ($R^{\mathrm{tag}}_c$), objects ($R^{\mathrm{obj}}_c$), attributes ($R^{\mathrm{att}}_c$), and predicates ($R^{\mathrm{pred}}_c$):
\begin{equation}
    \SBR = \frac{1}{|\mathcal{C}|}\sum_{c \in \mathcal{C}} \frac{1}{4}\left(R^{\mathrm{tag}}_c + R^{\mathrm{obj}}_c + R^{\mathrm{att}}_c + R^{\mathrm{pred}}_c\right)
    \label{eq:senben-r}
\end{equation}
$\SBP$ is defined analogously with precision.
Per-category $\text{F1}_{\mathrm{SB},c}$ is the harmonic mean of $R_{\mathrm{SB},c}$ and $P_{\mathrm{SB},c}$:
\begin{equation}
    \SBF = \frac{1}{|\mathcal{C}|}\sum_{c \in \mathcal{C}} \frac{2 \cdot R_{\mathrm{SB},c} \cdot P_{\mathrm{SB},c}}{R_{\mathrm{SB},c} + P_{\mathrm{SB},c}}
    \label{eq:senben-f1}
\end{equation}
The two-level macro-averaging ensures each category contributes equally regardless of frame count.
We prioritize recall because false negatives (missed sensitive content) are costlier than false positives in moderation.

Object matching uses the Hungarian algorithm on a class-aware IoU matrix with threshold IoU $\geq 0.5$.
Objects are ``sensitive'' if they are inherently sensitive (\eg, weapons, substances) or carry at least one SenBen attribute.
Predicate matching uses greedy triplet matching with domain-specific synonym sets and supports symmetric predicates (\eg, \emph{kissing}, \emph{holding}).
All four components evaluate only the SenBen-specific vocabulary (\SenBenObjects{} objects, \SenBenAttributes{} attributes, \SenBenPredicates{} predicates); non-sensitive elements are excluded from scoring.
Caption similarity (BGE-M3 cosine) is reported separately.

\subsection{Multi-Task Training}
\label{sec:training}

\begin{figure*}[t]
\centering
\includegraphics[width=0.8\textwidth]{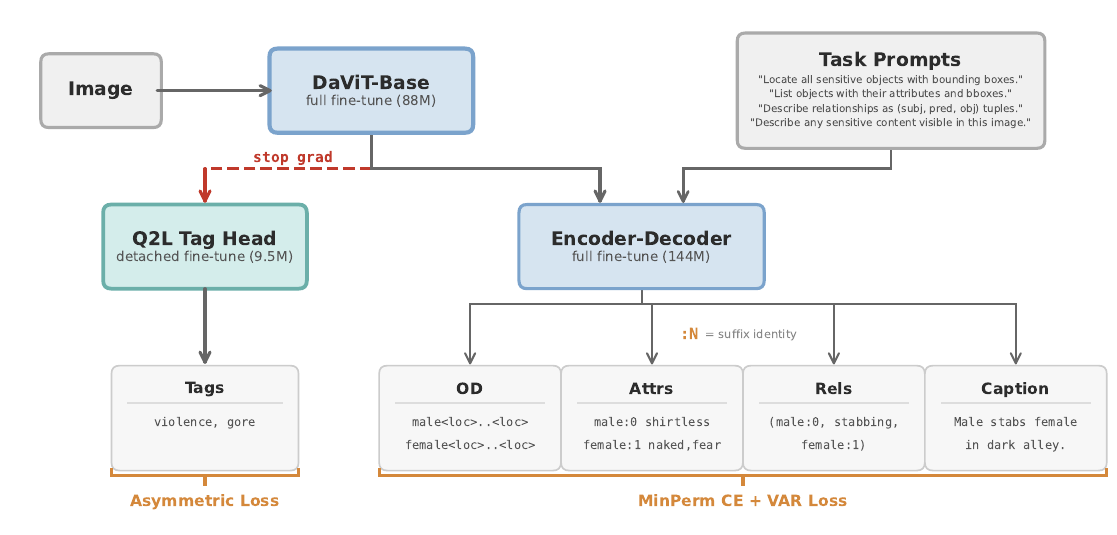}
\caption{Multi-task training architecture. DaViT vision features feed both a decoupled Q2L tag head (detached from the decoder) trained with Asymmetric Loss, and a fully fine-tuned encoder-decoder for four scene graph tasks. Suffix-based object identity (\textcolor{orange!80!black}{\textbf{:N}}) replaces bounding box tokens in attribute and predicate outputs. VAR loss (Eq.~\ref{eq:var}) penalizes low recall on sensitive vocabulary tokens across decoder tasks.}
\label{fig:pipeline}
\end{figure*}

We fully fine-tune \fl-base~\cite{xiao2024florence2} (231M) on five tasks, each triggered by a dedicated task token (Figure~\ref{fig:pipeline}).
An input frame is first encoded by the DaViT (Dual Attention Vision Transformer) vision encoder, whose features then feed two parallel pathways: (1)~a fully fine-tuned encoder-decoder that handles four scene graph tasks (object detection, attribute prediction, predicate prediction, and captioning) and (2)~a decoupled Query2Label (Q2L) tag head (+10M parameters) for multi-label sensitivity classification, bringing the total to 241M.
Inference uses beam search ($B{=}3$).


\noindent\textbf{Suffix-based object identity.}
In the attribute and predicate tasks, we replace bounding box tokens with \texttt{:N} identity suffixes assigned in raster-scan order (\eg, \texttt{male}, \texttt{female}, \texttt{male:1} for two males and one female).
Object detection retains \texttt{name<loc>} format.
This provides cross-task object consistency and reduces sequence length by 17--27\%.
The design is inspired by FactualSceneGraph~\cite{factual_scene_graph}.\\
\noindent\textbf{Vocabulary-Aware Recall (VAR) Loss.}
Sensitive vocabulary tokens (\eg, \emph{naked}, \emph{stabbing}, \emph{gore}) are rare relative to structural tokens (\eg, coordinates, punctuation) in the autoregressive sequence.
Standard cross-entropy treats all token positions equally, underweighting rare but critical terms.
We add a differentiable recall penalty:
\begin{equation}
    \mathcal{L} = \mathcal{L}_\text{CE} + \lambda \cdot (1 - R_s)^\gamma
    \label{eq:var}
\end{equation}
where $R_s = \frac{1}{|S|}\sum_{i \in S} p_i$ is the mean softmax probability at ground-truth token positions $S$ belonging to the predefined sensitive vocabulary, $\lambda$ weights the recall penalty relative to cross-entropy, and $\gamma$ is a focusing exponent that amplifies the penalty when sensitive token recall is low.
A BPE first-token filter identifies sensitive positions with ${\sim}20{\times}$ speedup over brute-force subsequence scanning.
VAR adapts Skeleton Recall Loss~\cite{skeleton_recall_loss} from pixel segmentation to autoregressive token generation (Table~\ref{tab:var-compare}).

\begin{table}[h]
\centering
\caption{VAR Loss in context of recall-focused losses.}
\label{tab:var-compare}
\small
\resizebox{\columnwidth}{!}{%
\begin{tabular}{@{}lccc@{}}
\toprule
Aspect & Recall Loss & Skeleton Recall & VAR (Ours) \\
\midrule
Domain & Segmentation & Segmentation & Autoregressive \\
Unit & Per-class pixels & Thin structures & Vocab tokens \\
Formulation & Multiplicative & Additive & Additive \\
\bottomrule
\end{tabular}}
\end{table}

\noindent\textbf{Q2L tag head.}
Inter-task gradient analysis reveals that tag classification gradients are nearly orthogonal to scene graph tasks in the decoder, while scene graph tasks form a tight cooperative cluster.
This motivates a fully decoupled tag head: a Query2Label~\cite{liu2021query2label} transformer decoder with 16 learnable label queries attending to DaViT encoder features, detached from the encoder-decoder to prevent co-adaptation.
The tag head uses Asymmetric Loss (ASL)~\cite{ridnik2021asymmetric} to handle class imbalance.
We evaluate two configurations: \emph{balanced} ($\gamma^-{=}4, \gamma^+{=}1$) and \emph{aggressive} ($\gamma^-{=}7, \gamma^+{=}0$), which trade off tag precision for higher recall.\\
\noindent\textbf{Supporting ingredients.}
MinPermutationCE addresses the arbitrary ordering problem in autoregressive set generation by selecting the ground-truth permutation that minimizes cross-entropy:
\begin{equation}
    \mathcal{L}_{\text{MinPerm}} = \min_{\pi \in \Pi(E)} \text{CE}\!\left(\mathbf{z},\; \text{enc}(\text{join}(\pi(E)))\right)
    \label{eq:minperm}
\end{equation}
where $E{=}\{e_1,\ldots,e_k\}$ is the set of unordered elements, $\Pi(E)$ enumerates valid permutations, and $\text{join}(\cdot)$ reconstructs the formatted output string.
Elements exceeding $k{=}5$ retain their original order ($<$2\% of frames).
Unlike OaXE~\cite{du2021oaxe}, MinPermCE operates on autoregressive models where permuting the target changes the causal conditioning chain; unlike $\sigma$-GPTs~\cite{pannatier2024sigmagpts}, it selects the best permutation rather than training on random ones.
Scheduled sampling~\cite{bengio2015scheduled} mixes ground-truth and model-predicted tokens ($p: 0{\rightarrow}0.3$ over 500 steps), reducing exposure bias during training.
Label smoothing ($\varepsilon{=}0.05$) regularizes the output distribution.

\subsection{Training Details}
\label{sec:details}

We fully fine-tune the Florence-2-base model using an effective batch size of 32 and a learning rate of $10^{-5}$. The model is trained for 15 epochs using the AdamW optimizer (weight decay: 0.01). We set a maximum sequence length of 256 tokens. For the multi-task objective, we apply task-specific weights: $2.0$ for tags and predicates, $1.5$ for object detection and attributes, and $1.0$ for captioning. For the VAR loss (\cref{eq:var}), we use hyperparameters $\lambda{=}0.1$, $\gamma{=}2$, with a linear warmup over the first 200 steps. The final model is selected based on the checkpoint achieving the highest $\SBR$ on the validation set.

\section{Experiments}
\label{sec:experiments}

We evaluate all models on the \senben test split, comprising \SenBenTestFrames{} frames from \SenBenTestMovies{} movies. For the VLM baselines, we employ the same structured scene graph prompt with zero-shot inference to ensure a direct and fair comparison.

\subsection{Ablation Study}
\label{sec:ablation}

Table~\ref{tab:ablation} details the incremental performance gains provided by each proposed component.
The final row adds the Q2L tag head on top of the best decoder configuration.

\begin{table}[t]
\centering
\caption{System ablation study. $\SBR$:~SenBen Recall, $\SBF$:~SenBen F1, $\text{F1}^{\mathrm{tag}}$:~macro tag F1 ($\SBR$/$\SBF$ category-macro-averaged).}
\label{tab:ablation}
\small
\begin{tabular}{@{}lccc@{}}
\toprule
System & $\SBR$ & $\SBF$ & $\text{F1}^{\mathrm{tag}}$ \\
\midrule
CE (baseline)       & .349 & .389 & .544 \\
+Suffix             & .366 & .406 & .509 \\
+VAR                & .376 & .408 & .509 \\
+MinPermCE          & .386 & .415 & .532 \\
+Label smooth.      & .385 & .419 & .533 \\
+Sched.\ samp.      & .392 & .422 & .516 \\
\midrule
+Q2L balanced       & \textbf{.413} & \textbf{.428} & \textbf{.594} \\
\bottomrule
\end{tabular}
\end{table}

Our final system achieves a $+6.4$ percentage point (pp) improvement in $\SBR$ and a $+3.9$pp increase in $\SBF$ over the standard CE baseline.

\noindent\textbf{Component Importance.} A leave-one-out analysis (Table~\ref{tab:loo-category}) ranks suffix-based object identity as the most critical component ($-3.8$pp~$\SBR$), followed by VAR loss  ($-3.4$pp).
\noindent\textbf{Category-Specific Impact.} While suffix identity is vital for grounding complex interactions in the violence ($-6.5$pp) and sexual ($-4.5$pp) categories, VAR Loss proves most effective at addressing the extreme vocabulary imbalance in the sexual ($-7.7$pp) and immodesty ($-5.7$pp) categories.
\noindent\textbf{Tag Precision.} While the decoder's recall optimization initially leads to a decline in tag F1, the introduction of the decoupled Q2L tag head reverses this trend, providing a significant $+7.8$pp boost in tag F1 compared to the full decoder configuration.
\begin{table}[t]
\centering
\caption{Per-category $\SBR$ change when removing each ingredient from the full decoder (VAR+SS+MinP+LS+Suffix). Values are $\Delta\SBR$ in percentage points.}
\label{tab:loo-category}
\small
\begin{tabular}{@{}lcccccc@{}}
\toprule
Removed & immod & sexual & viol & subst & other & avg \\
\midrule
$-$Suffix & $-4.6$ & $-4.5$ & $-6.5$ & $-2.9$ & $-0.4$ & $-3.8$ \\
$-$VAR    & $-5.7$ & $-7.7$ & $-2.0$ & $-0.9$ & $-0.9$ & $-3.4$ \\
$-$LS     & $-3.3$ & $-5.7$ & $+0.3$ & $+0.4$ & $+0.8$ & $-1.5$ \\
$-$SS     & $-1.6$ & $-2.8$ & $+0.3$ & $+0.8$ & $-0.4$ & $-0.7$ \\
$-$MinP   & $+0.5$ & $-2.4$ & $-1.4$ & $+1.7$ & $-0.4$ & $-0.4$ \\
\bottomrule
\end{tabular}
\end{table}

\subsection{Comparison with Baselines}
\label{sec:baselines}

\paragraph{VLM baselines.}
Table~\ref{tab:baselines} compares our models against frontier VLMs on full \senben metrics.
All VLMs use the same structured scene graph prompt with zero-shot inference.
Our model outperforms all VLMs except Gemini on object detection ($R^{\mathrm{obj}}$ $0.42$ \vs next best $0.30$) and caption similarity ($0.77$ \vs $0.65$).
Gemini~3~Pro achieves the highest $\SBF$ (0.647), partly because it generated the initial annotations that were refined into ground-truth labels, creating a stylistic advantage that should be considered when interpreting its scores.
Among GPT models, reasoning mode matters: GPT-5.2 with medium reasoning ($\SBF{=}0.362$) gains $+5.8$pp~$\SBF$ over GPT-5.2 without reasoning (0.304), mainly from better predicates ($+10.7$pp~$R^{\mathrm{pred}}$).
GLM-4.6V ($\SBF{=}0.364$) slightly edges GPT-5.2 medium reasoning, showing 10B open-weight models rival frontier API models on this task.
Both Claude models lack native bounding box grounding, explaining weak object detection ($R^{\mathrm{obj}}$ $0.03{-}0.08$) despite strong tag detection ($\text{F1}^{\mathrm{tag}}$ $0.64{-}0.66$).

\begin{table}[t]
\centering
\caption{\senben results on \SenBenTestFrames{} test frames. $\SBR$/$\SBF$:~SenBen Recall/F1, $\text{F1}^{\mathrm{tag}}$:~tag F1, $R^{\mathrm{obj}}$:~obj. recall, cap:~caption sim.}
\label{tab:baselines}
\small
\resizebox{\columnwidth}{!}{%
\begin{tabular}{@{}lrccccc@{}}
\toprule
Model & Params & $\text{F1}^{\mathrm{tag}}$ & $R^{\mathrm{obj}}$ & $\SBR$ & $\SBF$ & cap \\
\midrule
Gemini 3 Pro (low reas.)  & --- & .806 & .295 & .652 & .647 & .642 \\
Gemini 3 Flash (low reas.) & --- & .784 & .271 & .593 & .583 & .654 \\
\midrule
Q2L-bal (ours)     & 241M & .594 & \textbf{.420} & .413 & .428 & \textbf{.771} \\
Q2L-agg (ours)     & 241M & .457 & .409 & .449 & .431 & .772 \\
\midrule
Claude Opus 4.6    & --- & .658 & .082 & .327 & .404 & .598 \\
GLM-4.6V (reas.)   & 10.3B & .492 & .123 & .291 & .364 & .563 \\
GPT-5.2 (med.\ reas.) & --- & .608 & .072 & .319 & .362 & .616 \\
Qwen3-VL-8B        & 8.3B & .469 & .104 & .286 & .340 & .548 \\
Claude Sonnet 4.6  & --- & .643 & .034 & .277 & .339 & .590 \\
GPT-5-mini (med.\ reas.) & --- & .659 & .040 & .285 & .330 & .605 \\
GPT-5.2            & --- & .550 & .052 & .247 & .304 & .583 \\
\bottomrule
\end{tabular}}
\end{table}
\noindent\textbf{Safety classifiers.}
Table~\ref{tab:safety} compares tag detection only, since safety classifiers do not produce scene graphs.
No commercial API covers more than 8 of the 16 MECD tags.
Our Q2L-balanced model covers all 16 tags and achieves $\text{F1}^{\mathrm{tag}}{=}0.594$ \vs the best commercial API (Azure, $\text{F1}^{\mathrm{tag}}{=}0.430$ on 5 tags).
On binary safe/unsafe detection, it reaches $\text{F1}^{\mathrm{s}}{=}0.847$ \vs OpenAI Moderation at 0.664.
ShieldGemma~2 (4B) achieves only $\text{F1}^{\mathrm{tag}}{=}0.089$, illustrating the domain gap between AI-generated and real movie content.
Most existing safety classifiers were trained on explicit content that dominates the frame (pornographic images, overt violence) and operate at low resolution.
\senben movie frames present a harder challenge: sensitive content may appear in the background, occupy a small portion of the frame, or require contextual understanding (e.g., a partially obscured drug scene, implied nudity, a medical procedure behind foreground activity).
Narrow-scope classifiers (NudeNet, SD Safety Checker, LAION) cover 1--2 tags and only detect overt nudity/NSFW---they cannot recognize violence, substances, or implied sexuality, categories requiring scene-level reasoning rather than pixel-level pattern matching.
NudeNet's body-part detector, for instance, is a YOLOv8m~\cite{yolov8_ultralytics} model at 640px, which limits it to anatomical exposure cues.

\begin{table}[t]
\centering
\caption{Tag detection comparison. Tags:covered MECD tags. $\text{F1}^{\mathrm{tag}}$:~macro tag F1 over supported tags. $\text{F1}^{\mathrm{s}}$:~safe/unsafe F1.}
\label{tab:safety}
\small
\begin{tabular}{@{}lrccc@{}}
\toprule
Model & Params & Tags & $\text{F1}^{\mathrm{tag}}$ & $\text{F1}^{\mathrm{s}}$ \\
\midrule
Q2L-bal (ours)         & 241M & 16/16 & \textbf{.594} & \textbf{.847} \\
Q2L-agg (ours)         & 241M & 16/16 & .457 & .835 \\
\midrule
Azure Content Safety   & --- & 5/16 & .430 & .504 \\
OpenAI Moderation      & --- & 6/16 & .411 & .664 \\
LlavaGuard 1.2         & 7.0B & 6/16 & .384 & .583 \\
Google SafeSearch       & --- & 8/16 & .341 & .476 \\
ShieldGemma 2           & 4.0B & 4/16 & .089 & .161 \\
\midrule
SD Safety Checker       & 304M & 2/16 & .333 & .472 \\
LAION Safety Checker    & 1.0B & 2/16 & .225 & .357 \\
NudeNet Detector        & 25.9M & 1/16 & .238 & .238 \\
NudeNet Classifier      & 8.5M & 1/16 & .117 & .117 \\
\bottomrule
\end{tabular}
\end{table}

\subsection{Qualitative Results}
\label{sec:qualitative}


\begin{figure*}[t]
\centering
\begin{subfigure}[t]{0.32\textwidth}
\centering
\includegraphics[width=\textwidth, trim=0 50 280 0, clip]{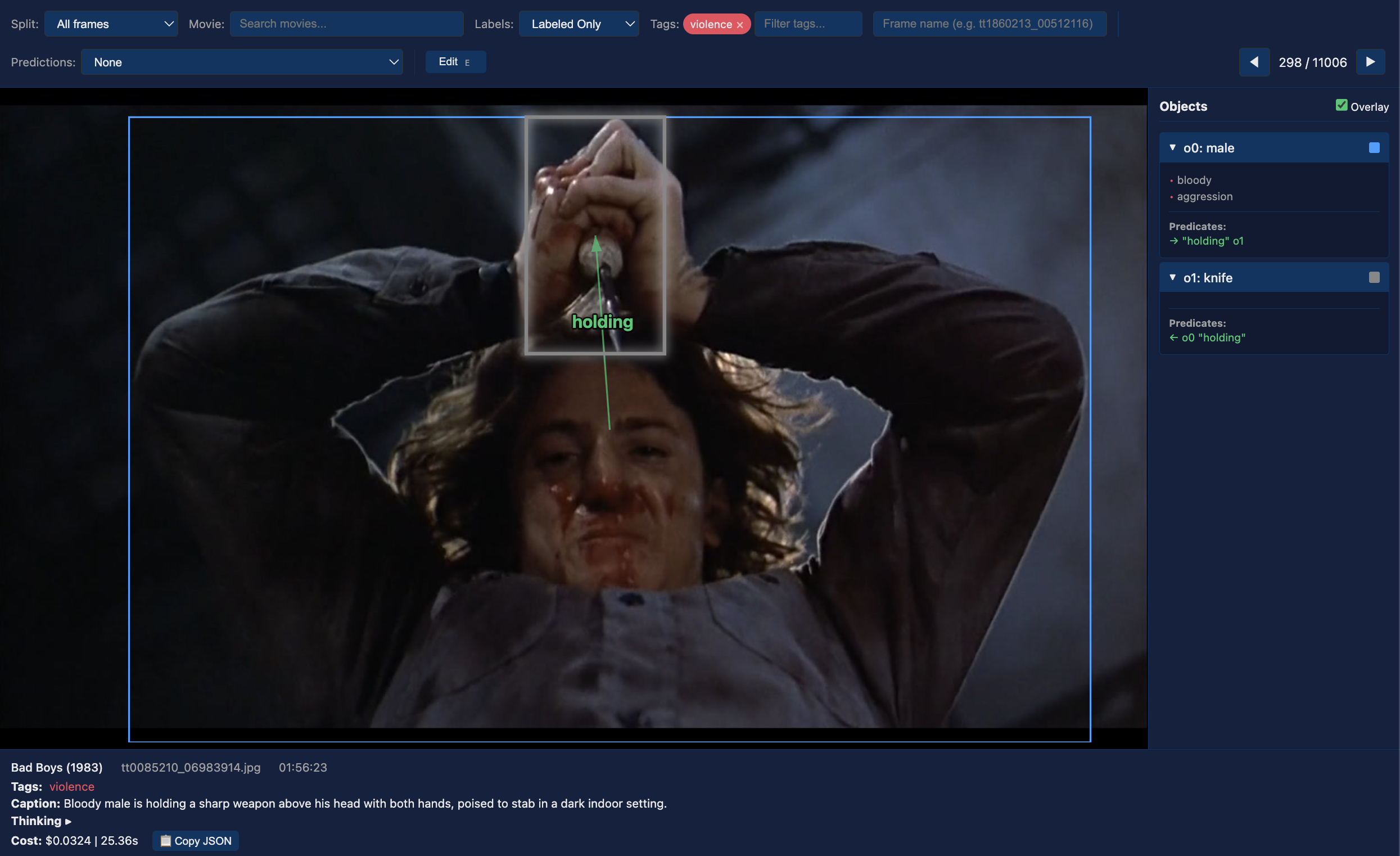}
\caption{\textbf{Violence (knife).} \texttt{male} (\emph{bloody, aggression}) $\xrightarrow{\text{holding}}$ \texttt{knife}. Tag: \texttt{violence}.}
\label{fig:qual-knife}
\end{subfigure}%
\hfill
\begin{subfigure}[t]{0.32\textwidth}
\centering
\includegraphics[width=\textwidth, trim=0 50 280 0, clip]{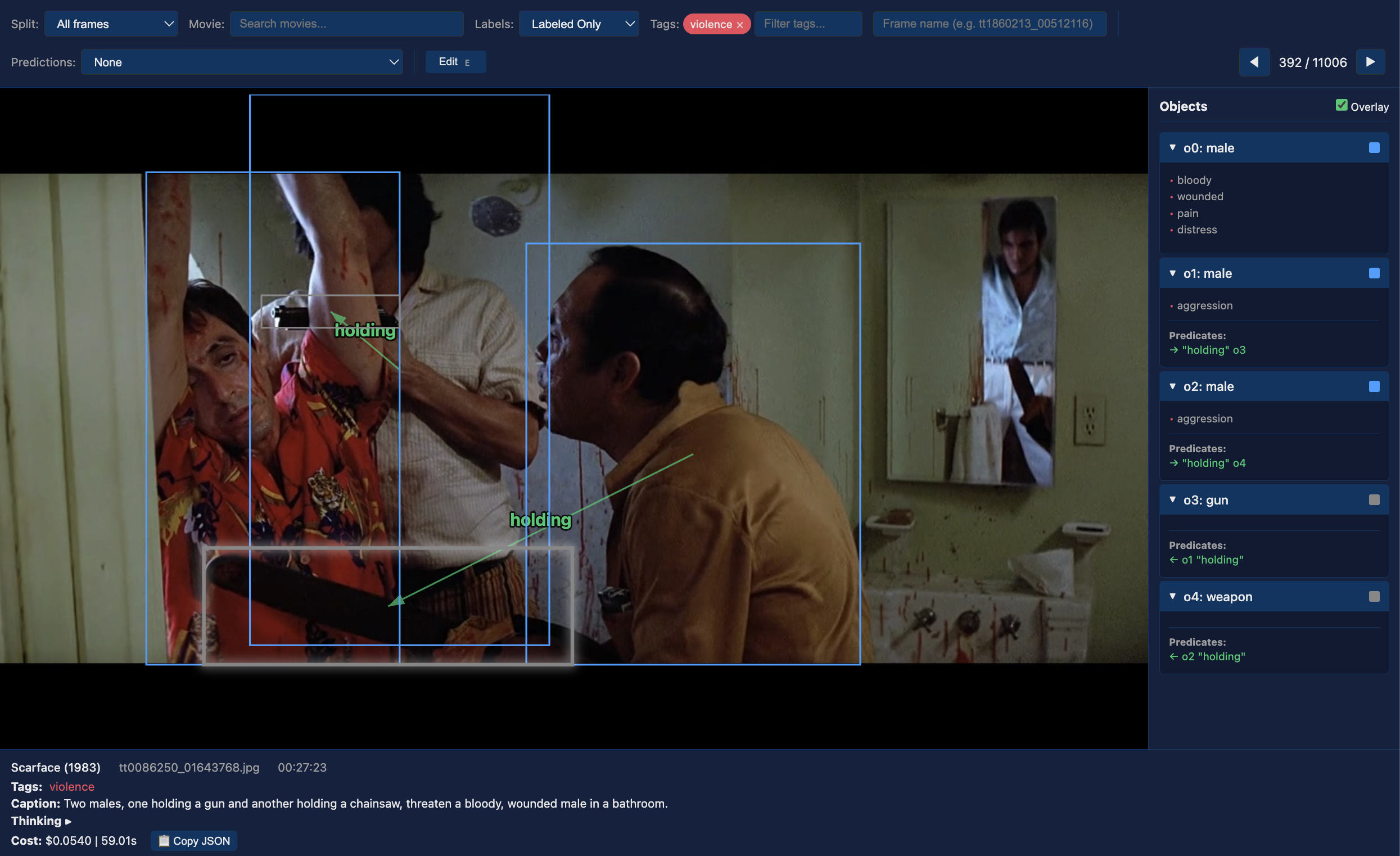}
\caption{\textbf{Violence (gun).} Three \texttt{male} objects (\emph{bloody, wounded, pain, aggression}) with \texttt{gun} and \texttt{weapon}. Tag: \texttt{violence}.}
\label{fig:qual-gun}
\end{subfigure}%
\hfill
\begin{subfigure}[t]{0.32\textwidth}
\centering
\includegraphics[width=\textwidth, trim=0 0 280 0, clip]{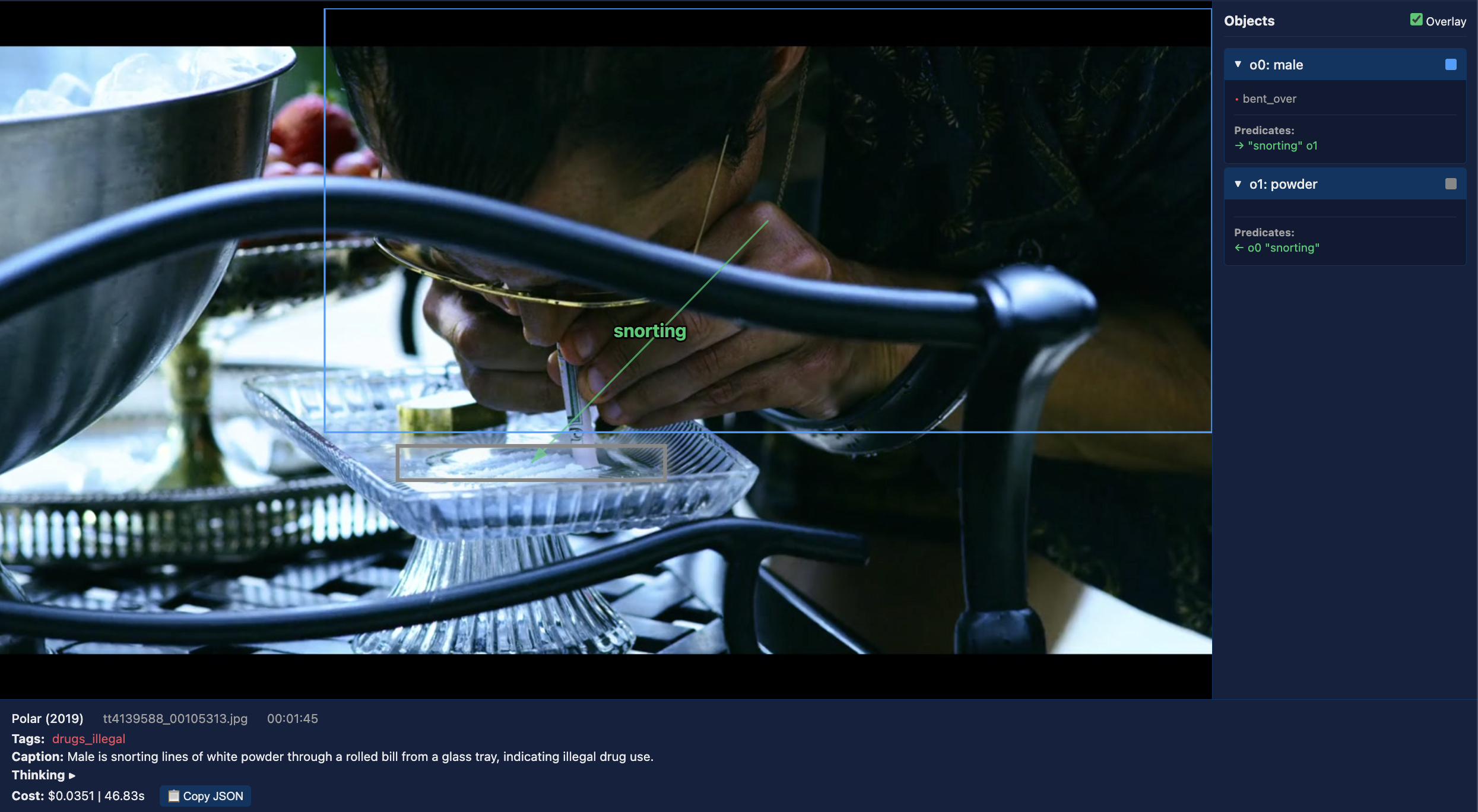}
\caption{\textbf{Substance use.} \texttt{male} (\emph{bent\_over}) $\xrightarrow{\text{snorting}}$ \texttt{powder}. Tag: \texttt{drugs\_illegal}.}
\label{fig:qual-drugs}
\end{subfigure}
\caption{Qualitative \senben annotations from our custom annotation web app. Each frame shows bounding boxes, predicate arrows, and the object panel with attributes. The scene graphs capture diverse sensitive content: single-object violence~(a), multi-actor violence~(b), and substance use~(c), all using the canonical SenBen vocabulary.}
\label{fig:qualitative}
\end{figure*}

Figure~\ref{fig:qualitative} shows three \senben annotations spanning two MECD categories.
The scene graphs range from a single subject--object pair (knife scene) to a multi-actor scenario with five objects and two predicates (gun scene).
Expression attributes (\emph{pain}, \emph{aggression}, \emph{distress}) and action predicates (\emph{holding}, \emph{snorting}) jointly determine the sensitivity tag, illustrating how \senben provides \emph{explainable} moderation decisions rather than opaque labels.

\subsection{Inference Efficiency}
\label{sec:efficiency}

Our model runs at 733\,ms/frame on an RTX~4090 in fp32 with beam search (B=3) across all five tasks, using only 1.2\,GB peak VRAM (5\% of the GPU).
This is $7.6\times$ faster than Qwen3-VL (5,614\,ms, 18.8\,GB) and $23\times$ faster than GLM-4.6V (17,056\,ms, 21.5\,GB), while requiring 16--18$\times$ less VRAM.
Compared to proprietary APIs, our model avoids per-frame costs entirely (\$0 \vs \$2--27 per 2K frames).
Table~\ref{tab:efficiency} reports full per-model latencies and costs.

\begin{table}[t]
\centering
\small
\caption{Inference efficiency. Latency:sequential 5-frame avg. VRAM:peak GPU memory. \$/2K:total API cost for 2000 frames.}
\label{tab:efficiency}
\footnotesize
\setlength{\tabcolsep}{4pt}
\begin{tabular}{@{}lrcrcc@{}}
\toprule
Model & Params & ms/fr & VRAM & \$/2K & $\SBF$ \\
\midrule
Q2L-bal (ours) & 241M & \textbf{733} & \textbf{1.2 GB} & \textbf{0} & .428 \\
\midrule
Claude Sonnet 4.6 & --- & 3438 & --- & 12.14 & .339 \\
Claude Opus 4.6 & --- & 4555 & --- & 20.02 & .404 \\
Gemini 3 Pro$^\dagger$ (low reas.) & --- & 5579 & --- & 26.58 & .647 \\
Qwen3-VL-8B & 8.3B & 5,614 & 18.8 GB & 0 & .340 \\
Gemini 3 Flash (low reas.) & --- & 6121 & --- & 5.80 & .583 \\
GPT-5.2 (med.\ reas.) & --- & 9019 & --- & 16.25 & .362 \\
GPT-5-mini (med.\ reas.) & --- & 13412 & --- & 4.49 & .330 \\
GLM-4.6V (reas.) & 10.3B & 17056 & 21.5 GB & 0 & .364 \\
\bottomrule
\end{tabular}
\begin{flushleft}
\scriptsize $^\dagger$Teacher model (generated initial labels, human-corrected).
\end{flushleft}
\end{table}

Our model's main weakness is predicate recall (0.24 \vs Gemini's 0.73); relationship reasoning is the hardest subtask for a compact model.
Its main strengths are object detection ($R^{\mathrm{obj}}$ $0.42$ \vs next-best VLM $0.30$) and captioning (0.77 \vs 0.65 cosine similarity).
Performance on the \emph{other} category shows high variance (4.3\% of test frames, min 11 samples/tag), suggesting future data augmentation.

\section{Conclusion}
\label{sec:conclusion}

We presented \senben, the first large-scale scene graph benchmark for sensitive content with person-level affective attributes that ground \emph{who} exhibits which behavioral state and \emph{what} triggers it, and a multi-task recipe that yields a compact model competitive with frontier VLMs on grounded metrics.
Suffix-based object identity and VAR Loss are the two most impactful ingredients, with category-specific effects: suffix is critical for violence and sexual content, while VAR primarily helps sexual and immodesty categories.
Task affinity analysis reveals that tag classification gradients are nearly orthogonal to scene graph tasks in the decoder, motivating the decoupled Q2L tag head that provides $+7.8$pp~$\text{F1}^{\mathrm{tag}}$.

\noindent\textbf{Limitations.} Gemini~3~Pro generated the initial annotations (subsequently human-corrected) and is also the strongest baseline, creating potential stylistic bias in evaluation.
Due to the sensitive nature of the content, only the first author reviewed and corrected labels; we lack a formal inter-annotator agreement study.
The dataset is predominantly Western movies (1982--2023) and targets image-level analysis; actions spanning multiple frames (e.g., fight sequences) may be only partially captured without temporal context.
The ``other'' category (4.3\% of test frames, minimum 11 samples per tag) exhibits high variance.

\noindent\textbf{Future Work.} Dataset expansion with bias-aware movie selection (targeting rare tags), label refinement through systematic review and formal inter-annotator agreement, per-category ASL tuning for the Q2L tag head, temporal scene graph generation leveraging shot-level context for multi-frame behavioral dynamics, cross-dataset evaluation on USD and UnsafeBench, and domain transfer to user-generated and AI-generated images.

\section*{Acknowledgments}
The participation of Fatih Cagatay Akyon in CVPR 2026 was supported by Ultralytics, and the participation of Alptekin Temizel by the EPAM T\"{u}rkiye AI Research Fund, administered by Graduate School of Informatics.

{
    \small
    \bibliographystyle{ieeenat_fullname}
    \bibliography{main}
}

\end{document}